\documentclass[10pt,twocolumn,letterpaper]{article}
\usepackage[accsupp]{axessibility}
\usepackage{cvpr}              

\usepackage{graphicx}
\usepackage{amsmath}
\usepackage{amssymb}
\usepackage{booktabs}
\usepackage{multirow}
\usepackage{tikz}
\usepackage{mathtools}
\usetikzlibrary{positioning}
\usetikzlibrary{arrows.meta}

\usepackage{overpic}
\definecolor{cvprblue}{rgb}{0.21,0.49,0.74}
\usepackage[pagebackref,breaklinks,colorlinks,allcolors=cvprblue,hyperfootnotes=false]{hyperref}

\def\paperID{40355} \def\confName{CVPR}
\def\confYear{2026}

\title{TESO: Online Tracking of Essential Matrix by Stochastic Optimization}

\author{Jaroslav Moravec\\
Czech Technical University\\
in Prague, Czech Republic\\
{\tt\small moravj34@fel.cvut.cz}
\and Radim \v S\'ara\\
Czech Technical University\\
in Prague, Czech Republic\\
{\tt\small sara@fel.cvut.cz}
\and
Akihiro Sugimoto\\
National Institute of Informatics\\
Tokyo, Japan\\
{\tt\small sugimoto@nii.ac.jp}
}

\begin{document}
\maketitle

\begingroup
\begin{tikzpicture}[remember picture, overlay]
      \node[draw, align=center] at (8.35, 6.25) {\Large{\textcolor{red}{This work was accepted for oral presentation at CVPR 2026.}}\\ \Large{ \textcolor{red}{Please cite the CVPR version once it is available.}}};
      \node[draw, align=left] at (8.25,-18.65) {\textcopyright 2026 IEEE. Personal use of this material is permitted. Permission from IEEE must be obtained for all other uses, in any current\\ or future media, including reprinting/republishing this material for advertising or promotional purposes, creating new collective\\ works, for resale or redistribution to servers or lists, or reuse of any copyrighted component of this work in other works.};
    \end{tikzpicture}
\renewcommand{\thefootnote}{} \footnotetext{Code available at \url{https://github.com/moravecj/teso}.}
\endgroup

\begin{abstract}
\noindent
Maintaining long-term accuracy of stereo camera calibration parameters is important for autonomous systems' perception. This work proposes Online \textbf{T}racking of \textbf{E}ssential Matrix by \textbf{S}tochastic \textbf{O}ptimization (TESO).
The core mechanisms of TESO are: 1) a~robust loss function based on kernel correlation over tentative correspondences, 2) an adaptive online stochastic optimization on the essential manifold. TESO has low CPU and memory requirements, relies on a~few hyperparameters, and eliminates the need for data-driven training, enabling the usage in resource-constrained online perception systems.

We evaluated the influence of TESO on geometric precision, rectification quality, and stereo depth consistency. On the large-scale MAN TruckScenes dataset, TESO tracks rotational calibration drift with 0.12$^\circ$ precision in the Y-axis--critical for stereo accuracy--while the X- and Z-axes are five times more precise. Tracking applied to sequences with simulated drift shows similar precision with respect to the reference as tracking applied to no-drift sequences, indicating the tracker is unbiased.
On the KITTI dataset, TESO revealed systematic inconsistencies in extrinsic parameters across stereo pairs, confirming previous published findings. We verified that intrinsic decalibration affected these errors, as evidenced by the conflicting behavior of the rectification and depth metrics. After correcting the reference calibration, TESO improved its rotation precision around the Y-axis 20$\times$ to 0.025$^\circ$ and its depth accuracy 50$\times$. Despite its lightweight design, direct optimization of the proposed TESO loss function alone achieves accuracy comparable to that of neural network–based single-frame methods.

\end{abstract} \section{Introduction}
\label{sec:intro}

\AtBeginEnvironment{overpic}{\footnotesize}
\begin{figure}[t]
    \centering
    \raisebox{3ex}{\includegraphics[scale=0.5]{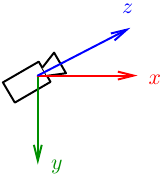}}
    \begin{overpic}[width=0.8\linewidth, trim={2cm 0 2.8cm 0},clip]{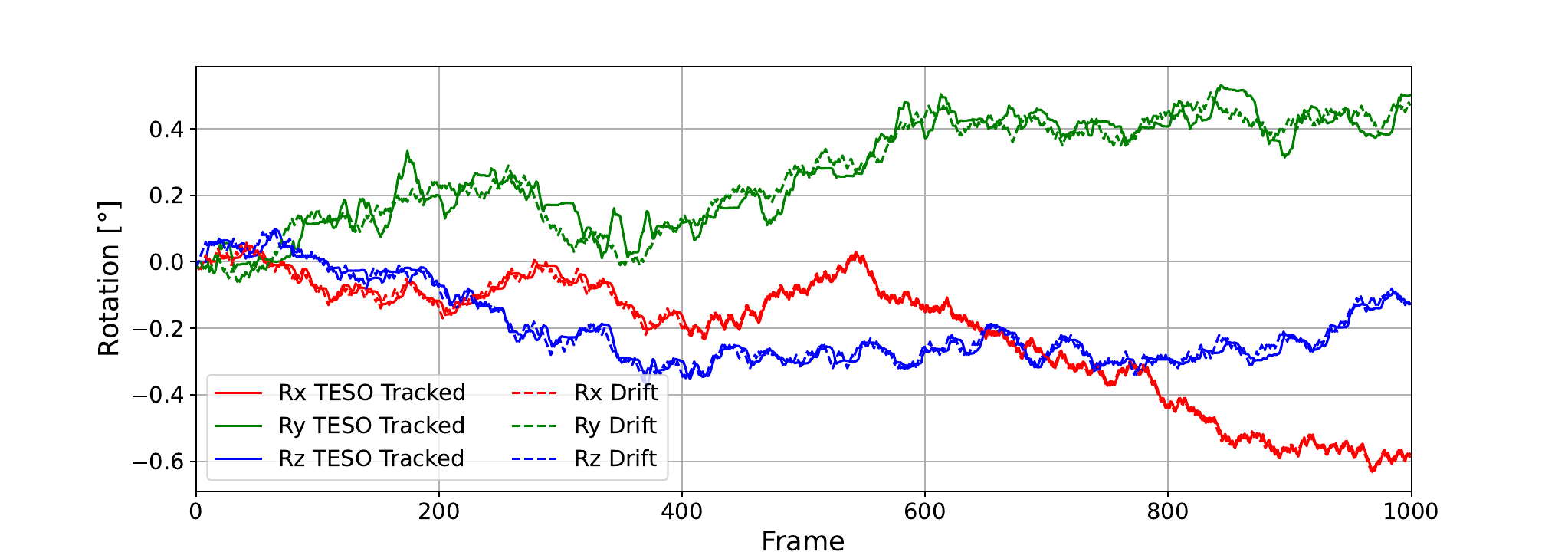}
        \put(0,-5){(a) Coordinate system and TESO tracking}
    \end{overpic}\vspace{1.5em}
\begin{overpic}[width=\linewidth]{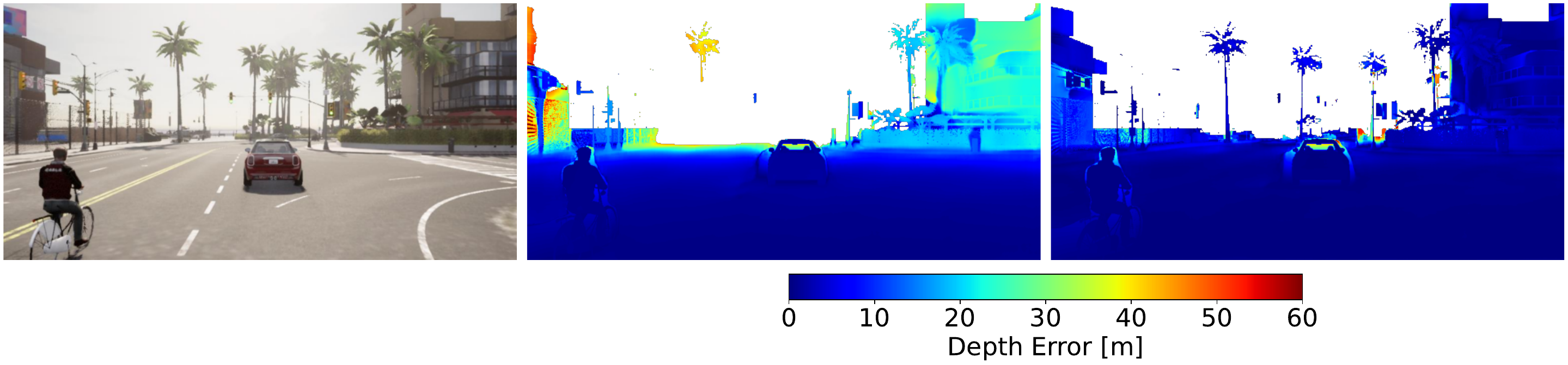}
        \put(10,-1.5){(b) Image}
        \put(38,-1.5){(c) Drifted}
        \put(80,-1.5){(d) TESO}
    \end{overpic}
    \caption{At the top (a) is an example of TESO tracking (solid) on a~sequence from the CARLA--Drift dataset with rotational calibration drift (dashed) of $\pm0.01^\circ$ per frame per DoF (see the coordinate system on the left). The easiest degrees to track are the rotations around X (red) and then Z (blue), which are easily observable in our kernelized epipolar error. The rotation around Y (green) is the most difficult to track, but also has the largest impact on depth estimation. At the bottom, there are stereo depth error maps for the image (b) on drifted calibration without tracking (c) and with TESO tracking (d). TESO can visibly reduce the effect of rotational drift because the disparity map is denser and the errors are lower.}
    \label{fig:carla_quant}
\end{figure}

The integration of heterogeneous sensors is a~core architectural principle of modern autonomous systems. Cameras, LiDARs, and radars provide complementary information about the environment, offering different temporal and spatial characteristics~\cite{waymo2019, Fent2024}. The effectiveness of such multi-sensor fusion -- and consequently of all downstream perception tasks -- depends critically on accurate geometric registration among sensors, i.e., their relative poses within a~common coordinate frame. Even small calibration errors can propagate through the perception pipeline, leading to inconsistent depth estimation, map misalignment, and, ultimately, reduced safety and reliability~\cite{dang2009, Cvisic21, Krishnan2025}.

In current practice, calibration is typically performed offline in controlled laboratory settings~\cite{zhang2000}. While such procedures can achieve high precision under ideal conditions, recalibrating deployed systems requires returning to specialized facilities, incurring downtime and operational costs. Therefore, maintaining accurate calibration during the operation of an autonomous system poses a~significant practical challenge. Mechanical vibrations, moving parts, thermal fluctuations, and material wear gradually alter the relative positions and orientations of sensors~\cite{Xiao2022}. These degradations are neither predictable nor uniform: they may appear as slow drifts, sudden shifts, or complex, environment-dependent patterns. Hence, we need an online tracking of the calibration parameters over time. To achieve short latency, small data batches must be used for parameter updates. This means that the process is sensitive to the variable information content in the sequence. Therefore, the challenge in online calibration parameter tracking is maintaining fast but stable convergence.

We address this challenge by proposing TESO, an adaptive online stochastic optimization method with a~robust loss function based on kernel correlation. In this work, we apply it to stereo vision systems, tracking the epipolar geometry parameters as they evolve on the essential matrix manifold. The essential matrix provides a~principled geometric framework for describing the epipolar geometry between stereo cameras, and its five degrees of freedom also encode both rotation and translation (up to baseline scale) between camera poses. By formulating calibration tracking as an optimization on the essential manifold, we preserve these geometric invariants while enabling adaptive parameter updates directly in their natural representation space. The kernel correlation introduces inherent robustness to outliers in tentative correspondences. This offers two key advantages. First, the choice of keypoint detector becomes non-critical (see Supplement \ref{seq_a:ablation}); even simple, fast-to-compute methods provide sufficient matches for most scenes. Second, the need for explicit outlier rejection via robust estimators is eliminated. Instead, we employ a~stochastic second-order optimization method with an adaptive learning rate. It can track a~non-stationary random process, and it naturally handles outliers due to the loss function design. Moreover, it is computationally efficient for online environments (autonomous vehicles, robotics systems, etc.). Our method requires only two hyperparameters and no data-driven training, making it straightforward to integrate into any stereo vision system. \cref{fig:overview} sketches the approach. Our main contributions are as follows:
\begin{itemize}
    \item We demonstrate that a~simple optimization of kernelized epipolar error can achieve competitive calibration performance with task- and data-specific neural network approaches.
    \item We show that combining rectification and depth metrics is important to capture all aspects of stereo calibration quality.
    \item With contradictory results of the two evaluation metrics, we found significant inconsistencies in the calibration parameters of the KITTI dataset, which corroborates the findings of other works. We also report that some stereo pairs contain a~larger decalibration than others.
    \item On the MAN TruckScenes dataset with a~wide baseline, camera vergence, and challenging scenes (highways, rain, etc.), TESO offered precise and unbiased tracking of a~fast-evolving simulated rotational drift of $\pm 0.01^\circ$ per frame per DoF.
    \item We introduce a~new dataset with a~synthetic camera orientation drift based on the CARLA simulator~\cite{Carla17}.
\end{itemize}

 \section{Related Work}
\paragraph*{Camera Pose Estimation.} These methods aim to determine the position and orientation of a~camera from a~single image with respect to a~given scene or potentially other images, making them closely related to the stereo extrinsic calibration discussed in this paper. However, methods for camera pose estimation are usually not built for online environments. Historically, such methods combined keypoint detectors (e.g., SIFT~\cite{lowe1999}) used in robust estimators (e.g., RANSAC~\cite{Fischler1981}) with solvers (e.g., the 5-point algorithm~\cite{nister04}). More recently, the field has shifted toward learning-based approaches: \Citet{Kendall15} introduced PoseNet, which directly regresses 6-DoF poses from monocular images using a~convolutional network. \Citet{chen21} factorized the 5D pose (up to scale) into several 3D direction vectors inducing spheres. Over these spheres, they regressed a~discrete distribution and estimated the expected value (i.e., the predicted direction vectors). Although these methods achieve strong generalization, the pose estimates are too coarse for stereo cameras. That is because they do not utilize the geometric consistency required for stereo calibration. More recently, \citet{rockwell2024} attempted to overcome this obstacle by combining a~learning-based approach with deep features~\cite{sun2021loftr} and robust estimation using a~solver.

\paragraph*{Online Stereo Calibration.} These approaches aim to maintain geometric alignment between stereo pairs as camera parameters evolve over time. Rather than relying mainly on pose errors, these methods also focus on rectification/epipolar errors. Dang et al.~\cite{dang2004, dang2006, dang2009} pioneered self-calibration approaches and gradually evolved their online tracking of stereo calibration parameters using an iterated extended Kalman Filter. They provided a~detailed analysis of the importance, sensitivity, and propagation to downstream error of each calibration parameter. They also evaluated the efficiency of three geometric constraints: reduced bundle adjustment, epipolar error, and trilinear constraints. More works have further extended these ideas; \Citet{hansen2012} proposed using epipolar constraints only, which removes the need for time-consuming temporal matching (such as bundle adjustment). \Citet{mueller2017} focused on calibration during urban driving, emphasizing stability in real-world scenarios. Specifically, they proposed using convolutional networks for segmentation to remove dynamic objects from the scene. \Citet{Cvisic22} proposed SOFT2, a~visual odometry system that continuously recalibrates the stereo extrinsics during operation using a~point-to-epipolar-line metric, effectively maintaining extrinsic consistency under real-world conditions. \Citet{kumar2024} proposed a~real-time learning-based rectification framework for stereo cameras trained on pose and optical flow errors. \Citet{gong2025} further studied online stereo rectification, using trained semi-dense matchers for precise feature extraction and RANSAC~\cite{Fischler1981} with the Levenberg-Marquardt algorithm to robustly find the rectification homographies (with~5 DoF).

Examining these relevant works, it is evident that stereo calibration and camera pose estimation are particularly challenging due to data variability and the low information content in some scenes. Existing approaches typically focus on selecting robust feature extractors~\cite{Cvisic22, kumar2024, gong2025} and/or estimators~\cite{rockwell2024, dexheimer2022, gong2025}. Our work adopts a~different perspective: We implicitly robustify the loss function itself through a~kernelized epipolar error, eliminating the need for complex outlier rejection schemes, robust estimators, or learned models. \section{Methods}
\label{sec:method}

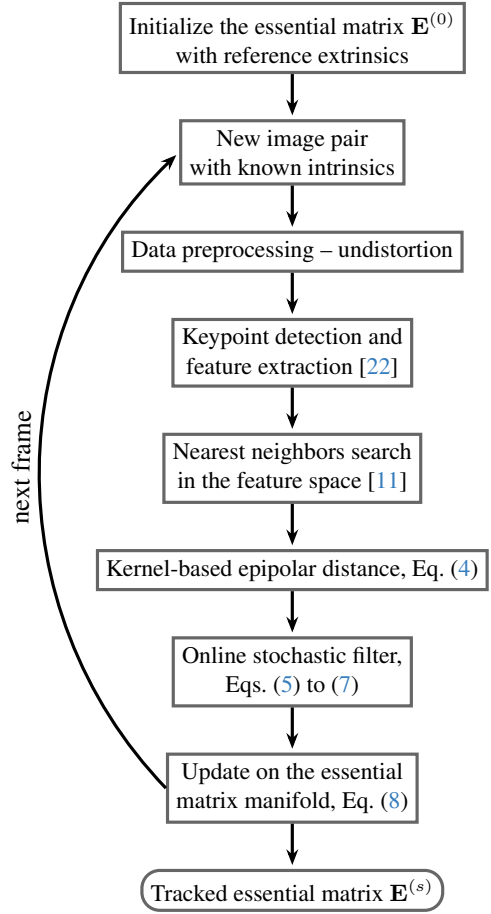
\begin{figure}
 \centering
 \def\lw{1.2pt}
\def\greendef{green}
\def\reddef{red}
\def\bluedef{blue}
\begin{tikzpicture}[node distance = .60cm,
roundnodegreen/.style={rectangle, line width=\lw, rounded corners=2ex, align=center,draw=\greendef!60, fill=\greendef!10, font = {\footnotesize}},
roundnode/.style={rectangle, line width=\lw, rounded corners=1.5ex, align=center, draw=black!60, font = {\small}},
roundnodered/.style={rectangle, line width=\lw, rounded corners=2ex, align=center, draw=\reddef!60, fill=\reddef!10, font = {\footnotesize}},
roundnodeblue/.style={rectangle, line width=\lw, rounded corners=2ex, align=center, draw=\bluedef!60, fill=\bluedef!10, font = {\footnotesize}},
squarednode/.style={rectangle, line width=\lw, align=center,draw=black!60, minimum size=5mm, font = {\small}},
squarednodeblue/.style={rectangle, line width=\lw, align=center,draw=\bluedef!60, minimum size=5mm, font = {\footnotesize}},
squarednodegreen/.style={rectangle, line width=\lw, align=center,draw=\greendef!60, minimum size=5mm, font = {\footnotesize}},
arrow/.style={->,line width=\lw,arrows={-Stealth[length=.6em,sep=1.4pt]}}
]

\node[squarednode] (A0) {Initialize the essential matrix $\mathbf{E}^{(0)}$\\ with reference extrinsics};
\node[squarednode] (A) [below=of A0] {New image pair\\with known intrinsics};
\node[squarednode] (B) [below=of A] {Data preprocessing -- undistortion};
\node[squarednode] (C) [below=of B] {Keypoint detection and\\ feature extraction \cite{lowe1999}};
\node[squarednode] (C1) [below=of C] {Nearest neighbors search\\in the feature space \cite{douze2024}};
\node[squarednode] (D) [below=of C1] {Kernel-based epipolar distance, \cref{eq:kernel_based_error}}; 
\node[squarednode] (E) [below=of D] {Online stochastic filter,\\ \cref{eq:ema,eq:memory,eq:sgd_step}};
\node[squarednode] (F) [below=of E] {Update on the essential\\ matrix manifold, \cref{eq:emm_update}};
\node[roundnode] (G) [below=of F] {Tracked essential matrix $\mathbf{E}^{(s)}$};
\draw[arrow] (A0.south) -- (A.north);
\draw[arrow] (A.south) -- (B.north);
\draw[arrow] (B.south) -- (C.north);
\draw[arrow] (C.south) -- (C1.north);
\draw[arrow] (C1.south) -- (D.north);
\draw[arrow] (D.south) -- (E.north);
\draw[arrow] (E.south) -- (F.north);
\draw[arrow] (F.west) to [bend left=45] node[midway, sloped, above, allow upside down]{next frame} (A.west);
\draw[arrow] (F.south) -- (G.north);
\end{tikzpicture}

  \caption{The full overview of TESO.}
\label{fig:overview}
\end{figure}
In this work, we propose a~\emph{calibration tracking} method for the essential matrix of a~stereo camera pair. The goal of such a~tracking procedure is to follow the temporal evolution of the essential matrix online during the operation of the autonomous system. Hence, it should meet the following requirements:
\begin{itemize}
    \item small computational overhead (CPU, GPU, memory),
    \item robustness to data (scene) variability and information content,
    \item small latency and real-time processing.
\end{itemize}

Our method utilizes online stochastic optimization with an adaptive learning rate~\cite{schaul2013} of the kernelized epipolar distance (similar to~\cite{Moravec24st}). This kernelization introduces uncertainty in each image-to-image match, thereby robustifying the method. There is no need to train a~specialized matcher for new data. Optimization is carried out right on the essential matrix manifold~\cite{helmke2007}. We now describe the method in more detail, follow \cref{fig:overview} for an outline.

\textbf{Data preprocessing.} Given a~stereo image pair $I_0^\text{orig}$ and $I_1^\text{orig}$, their camera matrices $\mathbf{K}_0$ and $\mathbf{K}_1$ and distortion parameters. We start by undistorting the image pair using the OpenCV~\cite{opencv_library} polynomial model, which yields a~new image pair $I_0$ and $I_1$.

\textbf{Keypoint detection and feature extraction.} Each image $I_j$ has a~set of $n^j$ detected keypoints $\{\mathbf{p}_i^j\in \mathbb{P}^{2}\}_{i=1}^{n^j}$ and a~set of their corresponding extracted descriptors $\{\mathbf{f}_i^j\in \mathbb{R}^{c}\}_{i=1}^{n^j}$. In this work, we use SIFT~\cite{lowe1999} to detect keypoints and extract descriptors ($c=128$) from images $I_0$ and $I_1$. The selection of the detector/descriptors is not a~critical aspect of this work (as discussed in Supplement \ref{seq_a:keypoint_detector}).

As we assume intrinsically calibrated cameras and look for an essential matrix, we also transform keypoints by their respective camera matrices $\mathbf{K}_j$ into the normalized space:
\[
    \mathbf{x}_i = \mathbf{K}_0^{-1} \mathbf{p}_i^0\quad\forall i,\quad \text{and}\quad \mathbf{y}_i = \mathbf{K}_1^{-1} \mathbf{p}_i^1\quad\forall i.
\]

We denote $\mathbf{X}=\{\mathbf{x}_i\}_{i=1}^{n^0}$ the set of normalized keypoints in the first image $I_0$ and $\mathbf{Y}=\{\mathbf{y}_i\}_{i=1}^{n^1}$ the set of normalized keypoints in the second image $I_1$.

\textbf{Essential matrix manifold.} Given a~correspondence $\mathbf{x} \sim \mathbf{y}$ between the images $I_0$ and $I_1$ with the relative rotation $\mathbf{R}\in \mathbb{R}^{3\times3}$ and translation $\mathbf{t}\in \mathbb{R}^3$, it must hold that
\begin{equation}\label{eq:epipolar_constraint}
    \mathbf{y}^\top\mathbf{E}\mathbf{x} = 0,
\end{equation}
where 
\begin{equation}\label{eq:E_fromRt}
    \mathbf{E} = [\mathbf{t}]_\times \mathbf{R}.
\end{equation}
Throughout this work, we assume the standard camera coordinate system, where the $z$-axis is the optical axis facing forward, the $x$-axis faces right, and the $y$-axis down~\cite{hartley2003} (as shown in \cref{fig:carla_quant}).

The change in baseline length has no effect on the constraint in \cref{eq:epipolar_constraint}, so only five degrees of freedom are observable in a~pair of images. We use optimization on the essential matrix manifold proposed in~\cite{helmke2007}. We start with the reference essential matrix $\mathbf{E}^{(0)} = \mathbf{E}^\text{ref}$ composed of the offline extrinsic calibration parameters $(\mathbf{R}, \mathbf{t})$ obtained, for example, from a~calibration room. In order to locally parameterize around the essential matrix $\mathbf{E}$, we decompose it using SVD into $\mathbf{E} = \mathbf{U}\mathbf{\Sigma}_0\mathbf{V}^{\top}$, normalized into $\mathbf{\Sigma}_0=\mathrm{diag}(1, 1, 0)$, and then use five parameters $\mathbf{\theta}\in \mathbb{R}^5$~\cite{helmke2007}:
\begin{equation}\label{eq:emm_map}
    \mathbf{E}(\mathbf{\theta}) = \mathbf{U}\,\text{expm}\left[\Omega_1(\mathbf{\theta})\right]\,\mathbf{\Sigma}_0\,\text{expm}\left[-\Omega_2(\mathbf{\theta})\right]\,\mathbf{V}^\top,
\end{equation}
where $\text{expm}$ is the matrix exponential and
\begin{equation*}
    \begin{split}
    \Omega_1(\mathbf{\theta})&=\frac1{\sqrt{2}}\begin{pmatrix} 0 & -\frac{\theta_3}{\sqrt{2}} & \theta_2\\ \frac{\theta_3}{\sqrt{2}} & 0 & -\theta_1\\ -\theta_2 & \theta_1 & 0 \end{pmatrix},\\
    \Omega_2(\mathbf{\theta})&=\frac1{\sqrt{2}}\begin{pmatrix} 0 & \frac{\theta_3}{\sqrt{2}} & \theta_5\\ -\frac{\theta_3}{\sqrt{2}} & 0 & -\theta_4\\ -\theta_5 & \theta_4 & 0 \end{pmatrix}.
\end{split}
\end{equation*}

\textbf{Kernel-based epipolar error.} As we only have tentative correspondences that contain many outliers, we use a~robust kernel correlation~\cite{tsin04}, which has been previously employed in LiDAR-Camera calibration~\cite{Moravec24} or stereo calibration monitoring~\cite{Moravec24st}. Given keypoints $\mathbf{X}\in I_0, \mathbf{Y}\in I_1$, we use the following loss function (lower is better)
\begin{equation}\label{eq:kernel_based_error}
  \begin{multlined}
    \mathcal{L}(\mathbf{\theta}\,|\,\mathbf{X}, \mathbf{Y}) = -\sum_{\mathbf{x}\in \mathbf{X}}\,\sum_{\mathbf{y}\in \text{NN}^1(\mathbf{x})}\text{exp}\left[-\frac{\left(\mathbf{y}^\top \mathbf{E}(\mathbf{\theta})\mathbf{x}\right)^2}{2\sigma^2}\right] - \\
     \sum_{\mathbf{y}\in \mathbf{Y}}\sum_{\mathbf{x}\in \text{NN}^0(\mathbf{y})}\text{exp}\left[-\frac{\left(\mathbf{y}^\top \mathbf{E}(\mathbf{\theta})\mathbf{x}\right)^2}{2\sigma^2}\right],
     \end{multlined}
\end{equation}
where $\text{NN}^1(\mathbf{x})$ are the $k$ most similar (in the descriptor space) keypoints $\mathbf{y}\in \mathbf{Y}$ to the keypoint $\mathbf{x}$. That is, the $k$ nearest neighbors of $\mathbf{f}^0_{\mathbf{x}}$ in $\mathbf{f}^1$. Similarly, $\text{NN}^0(\mathbf{y})$ are the $k$ most similar (in the descriptor space) keypoints $\mathbf{x}\in \mathbf{X}$ to the keypoint $\mathbf{y}$. We use~\cite{douze2024} to estimate kNN using the inner-product metric with $k=5$. The hyper-parameter $\sigma$ influences the width of the basin of attraction and the precision of the found parameters. We set it approximately to the pixel angular resolution (Vertical FoV/Width) of the camera. It is $\sigma=0.001$ for the CARLA--Drift and MAN TruckScenes datasets and $\sigma=0.00075$ for KITTI.

\textbf{Online stochastic optimization.} To minimize the loss function in \cref{eq:kernel_based_error}, we use an optimization method from \citet{schaul2013}: A~stochastic gradient descent method with an adaptive learning rate. It is based on the expected values of gradients $\mathbf{g}=\{g_i\}_{i=1}^5$, the diagonal of the Hessian matrix $\mathbf{h}=\{h_i\}_{i=1}^5$, and variance of gradients $\mathbf{v}=\{v_i\}_{i=1}^5$, per-parameter~$\mathbf{\theta}_i$:
\[g_i=\mathbb{E}\left[\frac{\partial \mathcal{L}}{\partial \theta_i}\right], \quad h_i=\mathbb{E}\left[\frac{\partial^2 \mathcal{L}}{\partial\theta_i^2}\right], \quad v_i=\mathbb{E}\left[\left(\frac{\partial \mathcal{L}}{\partial \theta_i}\right)^2\right].\]

The expectations are estimated by an exponential moving average with memory sizes $\mathbf{m}=\{m_i\}_{i=1}^5$, when frame $s$ arrives~\cite{schaul2013}:
\begin{equation}\label{eq:ema}
    g_i^{(s)} = \left(1 - \gamma_i^{(s)}\right) g_i^{(s-1)} + \gamma_i^{(s)}\frac{\partial \mathcal{L}}{\partial \theta_i}, \text{ with } \gamma_i^{(s)}=\frac{1}{m_i^{(s-1)}}
\end{equation}
and correspondingly for $\mathbf{h}$ and $\mathbf{v}$.

The memory size $m_i$ is also adaptively updated as
\begin{equation}\label{eq:memory}m_i^{(s)} = \left(1 - \frac{\left(g_i^{(s)}\right)^2}{v_i^{(s)} + \varepsilon}\right) \cdot  m_i^{(s-1)} + 1,\end{equation}
which increases the memory size if the squared gradients are lower than the variance (this stabilizes the optimization, when the random process changes abruptly). If the gradient is comparable to the variance it speeds the convergence up by lowering the memory size. The $\varepsilon=10^{-7}$ is a~small constant that prevents collapse when $v_i\rightarrow0$. We start with $m^{(0)}_i=1$ and accumulate the filtered values $g_i, h_i$ and $v_i$ for a~burn-in period of ten frames (the memory size is increased by one: $m_i^{(s)} = m_i^{(s-1)} + 1$, and no update is performed on the manifold).

After the burn-in period, the update on the essential manifold at step $s$ is estimated as~\cite{schaul2013}
\begin{equation}\label{eq:sgd_step}
    \Delta \theta_i^{(s)} = -\nu_i\,\,\frac1{h_i^{(s)}}\,\,\frac{\partial \mathcal{L}}{\partial \theta_i}\quad\text{with}\quad \nu_i = \frac{\left(g_i^{(s)}\right)^2}{v_i^{(s)} + \varepsilon}.
\end{equation}
The reasoning behind the adaptive learning rate term $\nu_i$ is similar to that in the memory size update above: It balances between 
quasi-Newton steps with $-\frac1{h_i^{(s)}}\,\,\frac{\partial \mathcal{L}}{\partial \theta_i}$ and gradient descend when the squared gradients decrease, reducing the size of the update and increasing convergence stability.

We use the mapping from \cref{eq:emm_map}, to update the essential manifold:
\begin{equation}
  \begin{split}
    \mathbf{U^{(s)}} = &\mathbf{U}^{(s-1)}\,\text{expm}\left[\Omega_1(\mathbf{\Delta\theta^{(s)}})\right] \text{ and}\\ {\mathbf{V}^{(s)}} = &{\mathbf{V}^{(s-1)}}\text{ expm}\left[\Omega_2(\mathbf{\Delta\theta^{(s)}})\right] \text{, which gives}\\ \mathbf{E}^{(s)} = &\mathbf{U}^{(s)}\mathbf{\Sigma}_0{\mathbf{V}^{(s)}}^{\top}.
    \end{split}\label{eq:emm_update}
\end{equation} \section{Datasets}
\paragraph*{CARLA--Drift.} As there is currently no real dataset with known decalibrations and ground truths, we have created one using the CARLA 0.9.15 simulator~\cite{Carla17}. We used the default pre-created TOWN10 map with 100 random autopiloted vehicles. From each of the 155 spawn points, we recorded an autopiloted drive for 1000 frames with a~simulated drift in the right camera's orientation of $\pm 0.01^\circ$ per frame per DoF. Both cameras have a~resolution of 1024$\times$ 512 \,px, a~vertical field of view of 70$^\circ$, and have a~baseline length of one meter. To assess the quality of stereo depth estimation, we added a~ground truth depth map camera with the same parameters and location as the left camera.

\paragraph*{KITTI} is the gold standard dataset for autonomous driving~\cite{Geiger2013}. In our paper, we use combinations of all four PointGray Flea2 cameras with their original undistorted and unrectified resolution of 1392$\times$512\,px. The horizontal field of view is around 70$^\circ$ and vertical 30$^\circ$. To assess the quality of the stereo depth estimation, we also utilize the LiDAR -- Velodyne HDL-64E, which features 64 layers and a~26.8$^\circ$ vertical field of view. Following the testing scheme from \citet{kumar2024}, we focus on evaluating our method using sequences from October 3, 2011 (27, 34, 42, 47 and 58).

\paragraph*{MAN TruckScenes} dataset~\cite{Fent2024} is the first large-scale multimodal dataset designed for autonomous trucking applications. It comprises 747 scenes (149 for testing) covering various environmental conditions and scenarios typical of commercial long-haul trucking (mainly from highway). The truck sensor suite includes four Sekonix SF3324 cameras, each with 1928$\times$1208 resolution and a~120$^\circ$ by 73$^\circ$ field of view, operating at a~10\,Hz sampling rate. We focus on the front left and right cameras, which have a~baseline of more than 2\,m. These two cameras have the largest common field of view, with about 14$^\circ$ vergence. To assess the quality of stereo depth estimation, we use the front left LiDAR -- Hesai Pandar64, which has 64 layers and a~40$^\circ$ vertical field of view.

\paragraph*{CARLA--FlowGuided.} Recently, in~\cite{kumar2024}, the authors introduced a~synthetic dataset based on the CARLA driving simulator, designed to evaluate online stereo rectification algorithms under challenging and diverse conditions. It features three RGB cameras, separated by a~baseline of 0.8 meters, capturing high-resolution images with a~2560$\times$1440 resolution and a~45$^\circ$ field of view. The simulated environment encompasses a~range of diverse urban and highway scenes, with extensive variation in lighting (from dawn to dusk) and weather (including fog, rain, overcast, and sunny conditions). The dataset incorporates artificial perturbations of the extrinsic parameters of the stereo camera, specifically random rotations on all three axes within $\pm$1 degree, while keeping the translations fixed. In total, it comprises 7,722 stereo pairs (774 for validation) with known ground truth camera poses and intrinsics. There is no LiDAR data. \section{Experiments}
First, we report the standard geometric precision estimated as the mean absolute error over all frames and sequences. Rotation and translation were received from the tracked essential matrix using OpenCV~\cite{opencv_library} decomposition. The translation is rescaled to the reference baseline length to compare (MAE) with GT values (see \cref{tab:geometric_precision_carla_drift} and \cref{tab:geometric_precision_man}). Alternatively, we compare only the angle between the found translation direction and GT translation (see \cref{tab:geometric_precision_carlafg}) as in~\cite{gong2025}. Second, we report the rectification metrics -- keypoint offset (KO) and vertical optical flow offset (VOF) -- introduced in~\cite{kumar2024}. For KO, we use SuperGlue matches~\cite{sarlin2020} and for VOF RAFT~\cite{teed2020}. We evaluate the images in grayscale and at a~resolution of $1024\times 512$ ($1241\times 376$ for KITTI) using the OpenCV rectification~\cite{opencv_library}. Finally, for KITTI and MAN TruckScenes, we use LiDAR to estimate the depth consistency (DC), measured by mean absolute error between the LiDAR-projection depth and the stereo depth computed from the RAFT horizontal optical flow. For CARLA--Drift, we use ground-truth depth maps instead of LiDAR projections. And since the CARLA-FlowGuided dataset does not contain LiDAR data or ground-truth depth maps, we use the method of \citet{kumar2024} to evaluate depth consistency, i.e., by comparing depth maps generated from ground-truth calibration parameters and our tracked ones. For KITTI only (see \cref{sec:kitti_exp}), we use the weighted average of all metrics over the five sequences based on the number of frames in each sequence. To make the tables (\cref{tab:stereo_metrics_carla_drift}, \cref{tab:stereo_metrics_kitti}, and \cref{tab:stereo_metrics_man}) more readable, we report the improvement (denoted as \verb|-I|) of the stereo metrics with our recalibration over the ground truth parameters. That is, a~negative number indicates that TESO recalibration is better than the reference calibration. In all metrics reported, a~lower number means better.

\subsection{Drift Tracking on CARLA--Drift}
\begin{table}[]
    \begin{subtable}[h!]{0.5\textwidth}
    \centering
    \begin{tabular}{c|ccc}
        Tracking & R$_\text{x}$ [$^\circ$] & R$_\text{y}$ [$^\circ$] & R$_\text{z}$ [$^\circ$] \\\hline
         Ours (TESO) & $\textbf{0.011}$ & $0.039$ & $\textbf{0.015}$ \\\vspace{1em}
         w/o & $0.157$ & $0.166$ & $0.175$ \\
          & T$_\text{x}$ [mm] & T$_\text{y}$ [mm] & T$_\text{z}$ [mm] \\\hline
          Ours (TESO) & $0.011$ & $2.413$ & $2.430$ \\
          w/o & $0$ & $0$ & $0$ \\
    \end{tabular}
\subcaption{Geometric Precision (lower is better)}
\label{tab:geometric_precision_carla_drift}
\end{subtable}\vspace{1em}
    \begin{subtable}[h!]{0.5\textwidth}
    \centering
    \begin{tabular}{c|ccc}
         Tracking & KO-I [px] & VOF-I [px] & DC-I [m] \\\hline
         Ours (TESO) & \textbf{0.03} & \textbf{0.07} & 0.52 \\
         w/o & 1.94 & 1.65 & 2.06\\
    \end{tabular}
\subcaption{Stereo Metrics (lower is better)}
\label{tab:stereo_metrics_carla_drift}
\end{subtable}
\caption{TESO performance on CARLA--Drift dataset, visualised as geometric precision \ref{tab:geometric_precision_carla_drift} and stereo metrics \ref{tab:stereo_metrics_carla_drift} (first rows) vs. metrics without tracking. One can see that TESO tracking consistently improves both the rotation precision and stereo metrics. The hardest DoF is R$_\text{y}$, which also has a~large negative impact on the depth consistency improvement (DC-I). Still, the drift is substantially mitigated by TESO, even in these two metrics -- geometric precision of R$_\text{y}$ and DC-I. }
\label{tab:carla_drift_results}
\hfill
\end{table}

This experiment shows performance under strictly controlled conditions, including known reference calibration. An example of TESO tracking on one of the 155 sequences is shown in \cref{fig:carla_quant}. See \cref{tab:carla_drift_results} for aggregated TESO tracking results on our CARLA--Drift dataset. The precision in rotations around X and Z is under $0.02^\circ$ on average, which is a~very good result considering the large drift magnitude of $\pm0.01^\circ$ per frame per DoF and the delay of the stochastic filter. The rotation around Y is the most difficult to observe in the epipolar error (used by TESO), but the obtained results of $0.039^\circ$ are still quite precise, considering the large vertical field of view and the low image resolution ($0.068^\circ$ per pixel). Even though we track all five parameters of epipolar geometry, only rotations drift. However, they are not compensated for by translation estimates, as evidenced by their millimeter-level precision. Stereo metrics have also been considerably improved by TESO (lower is better), specifically, the rectification-based KO-I and VOF-I. In terms of depth consistency, the improvement is not that large (still about four times lower, from 2.06 to 0.52), mainly due to the low observability of rotation around the Y-axis.

\subsection{Stereo Calibration Assessment on KITTI}\label{sec:kitti_exp}

This experiment discusses instances of decalibration previously observed by other authors. As shown in~\cite{Cvisic22}, it is possible to substantially improve the downstream results on KITTI data by recalibrating the stereo camera pair. With their new recalibrated intrinsic and extrinsic parameters~\cite{Cvisic21}, the authors were even able to surpass the results of LiDAR-based SotA methods in the odometry estimation task. We hypothesize that if the original KITTI calibration parameters~\cite{Geiger2013} are actually inconsistent, TESO should be able to observe these errors as well.

The results of the geometric precision of TESO on four different stereo pairs are shown in \cref{tab:geometric_precision_kitti}. When using the original KITTI intrinsic parameters from~\cite{Geiger2013}, the precision uncovers an interesting inconsistency in all tested pairs. The largest is between the pair 00-01, followed by 00-03 and 02-01, and the most precise one (according to TESO) is the color pair 02-03. When using the suggested offline intrinsic recalibration from~\cite{Cvisic21} (only available for the first pair 00-01), we can see that the large error (mainly in rotation around the Y axis, shown in red) of the pair disappears (shown in bold). From the results of the stereo metrics in \cref{tab:stereo_metrics_kitti}, we can draw similar conclusions. With the original parameters~\cite{Geiger2013}, TESO massively improves rectification-based errors (keypoint offset KO-I and vertical optical flow VOF-I), but it worsens depth consistency DC-I. This suggests that the error is present not only in the extrinsic parameters but also in the intrinsics. One can see that when using the recalibrated intrinsics from~\cite{Cvisic21}, the improvement in KO and VOF is smaller; however, depth consistency improves from 2\,m to 4\,cm.

\begin{figure}[t]
    \centering
    \includegraphics[width=0.95\linewidth]{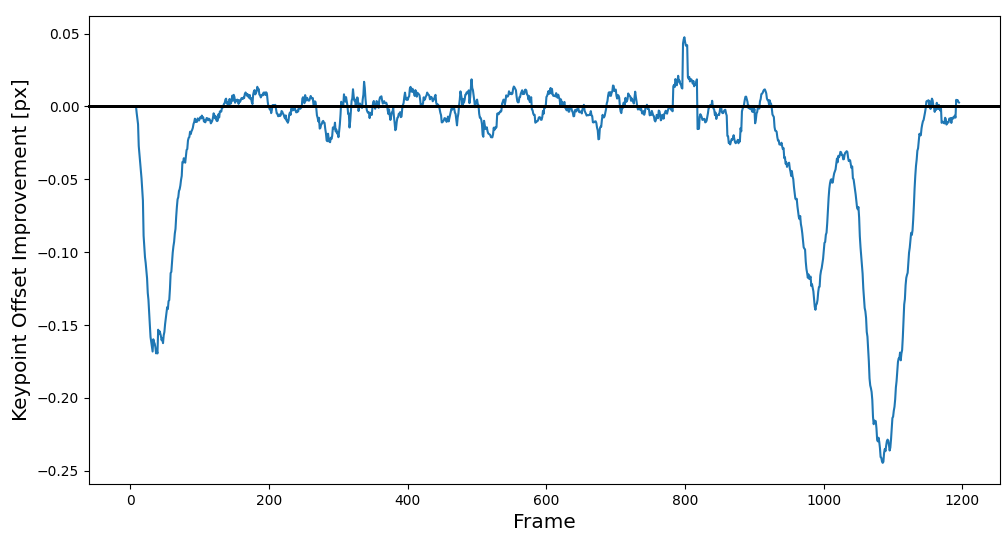}
    \caption{Keypoint offset improvement metric (negative values are improvements over the reference) evaluation on seq. 01 (drive \#42, 2011/10/03) with offline recalibration suggested in~\cite{Cvisic21}. TESO clearly improves this metric at the beginning and end of the sequence. This corroborates the findings from~\cite{Moravec24,Cvisic22, Brossard20}.}
    \label{fig:kitti_01_ko}
\end{figure}

In~\cite{Cvisic22}, they noticed that even with their offline recalibration from~\cite{Cvisic21}, the sequence \#42 from 2011/10/03 (Seq. 01 in KITTI Odometry Dataset) contains some sort of rotational decalibration in the beginning and end of the sequence (when the vehicle goes through a~turn). We confirm this observation: The evaluation of the keypoint offset improvement metric using TESO tracking on this sequence is shown in \cref{fig:kitti_01_ko}. The improvement is visible in the same parts of the sequence as in the stereo recalibration~\cite{Cvisic22}, LiDAR-Camera recalibration~\cite{Moravec24}, and IMU covariance estimation~\cite{Brossard20}.
\begin{table}[]
    \begin{subtable}[h!]{0.5\textwidth}
    \centering
    \begin{tabular}{cc|ccc}
         Stereo & Calib. & R$_\text{x}$ [$^\circ$] & R$_\text{y}$ [$^\circ$] & R$_\text{z}$ [$^\circ$] \\\hline
         \multirow{2}{*}{00-01} & \cite{Geiger2013} & 0.011 & \textcolor{red}{0.489} & 0.023 \\
          & \cite{Cvisic21} & 0.005 & \textbf{0.025} & \textbf{0.004} \\ \hline
         00-03 & \cite{Geiger2013} & 0.004 & 0.303 & 0.016 \\
         02-01 & \cite{Geiger2013} & 0.009 & 0.308 & 0.024 \\
         02-03 & \cite{Geiger2013} & 0.003 & 0.116 & 0.015 \\
    \end{tabular}
\subcaption{Geometric Precision (lower is better)}
\label{tab:geometric_precision_kitti}
\end{subtable}\vspace{1em}
    \begin{subtable}[h!]{0.5\textwidth}
    \centering
    \begin{tabular}{cc|ccc}
         Stereo & Calib. & KO-I [px] & VOF-I [px] & DC-I [m] \\\hline
         \multirow{2}{*}{00-01} & \cite{Geiger2013} & $-$0.121 & $-$0.169 & \textcolor{red}{2.14} \\
          & \cite{Cvisic21} & $-$0.008 & $-$0.016 & \textbf{0.04} \\ \hline
         00-03 & \cite{Geiger2013} & $-$0.035 & $-$0.100 & 1.40 \\
         02-01 & \cite{Geiger2013} & $-$0.054 & $-$0.081 & 1.15 \\
         02-03 & \cite{Geiger2013} & $-$0.010 & $-$0.023 & 0.35 \\
    \end{tabular}
\subcaption{Stereo Metrics (lower is better, negative means improvement over the reference parameters)}
\label{tab:stereo_metrics_kitti}
\end{subtable}
\caption{TESO performance on the KITTI dataset, visualised as geometric precision \ref{tab:geometric_precision_kitti} and stereo metrics \ref{tab:stereo_metrics_kitti}. With the original KITTI calibration parameters from~\cite{Geiger2013}, there is a~significant tracked decalibration in rotation around the Y-axis (middle column \ref{tab:geometric_precision_kitti}, red), which hurts the depth consistency of the stereo with respect to LiDAR (last column \ref{tab:stereo_metrics_kitti}, red). This occurs, despite improvements in both the keypoint offset and vertical optical flow metrics (negative in \ref{tab:stereo_metrics_kitti}, 1st row). When using the intrinsics from~\cite{Cvisic21} on stereo pair 00-01, both the geometric precision and the depth consistency errors decrease, indicating that decalibrations were also present in the intrinsics. The last three rows suggest that the intrinsic decalibrations were present in other sequences as well.}
\label{tab:kitti_results}
\hfill
\end{table}
\subsection{Drift Tracking on MAN TruckScenes}
This experiment shows tracking results on a~challenging real dataset. TESO performance on the MAN TruckScenes dataset is shown in \cref{tab:man_results}. Each sequence was repeated periodically ten times to obtain approximately 1600 frames in total per sequence. We first discuss the results on calibrated data (see 1st row \emph{Calibrated}), which show very good precision in rotations around X and Z, as well as translations in X (under 1\,mm) and Y. The most problematic rotation around $Y$ and translation in Z (which seems to compensate for each other on this dataset) still shows reasonable precision, when taking into account the large vergence (14$^\circ$), a~quite low pixel angular resolution of 0.06$^\circ$, and challenging conditions (highways, repetitive and distant scenes, variable weather conditions, etc.). The stereo metrics (see \cref{tab:stereo_metrics_man}), which were evaluated only on the last epoch (i.e., about 160 frames), also behave much better than in the case of the original calibration in KITTI \cref{tab:stereo_metrics_kitti}. For example, the depth consistency, although still quite high (with 0.166 meters on average), is much closer to zero. This suggests that the calibration parameters provided in the MAN TruckScenes dataset are of good quality. Note that the distances in the mostly highway scenes are generally greater than in urban scenes in other datasets.  Hence, the sequences are quite challenging for online calibration purposes. A~detailed analysis of TESO tracking on MAN TruckScenes, broken down by location, time of day, and weather, can be found in Supplement \ref{seq_a:man}.

We have also tested introducing a~synthetic drift of $\pm0.01^\circ$ per frame in each rotational degree of freedom individually (as in CARLA--Drift). This provides a~realistic decalibration that can occur during a~fast drive on highways on systems such as trucks with sensors on moving parts (the cabin). The results can be seen in \cref{tab:man_results} (the \emph{Drift} rows). The drop in precision is as expected, i.e., comparable to the drift amplitude of $\pm0.01^\circ$. Hence, TESO successfully tracked the synthetic drift in geometric precision and stereo metrics.

When working with the dataset, we noticed that a~slight change in focal length can result in significantly more precise calibration tracking of TESO in certain sequences. However, we were unable to consistently achieve better results with focal length recalibration compared to the results presented in \cref{tab:geometric_precision_man}. As a~potential direction for future research on the MAN TruckScenes dataset, we suggest incorporating focal length tracking in addition to essential matrix tracking.
\begin{table}[]
    \begin{subtable}[h!]{0.5\textwidth}
    \centering
    \begin{tabular}{c|ccc}
          Data & R$_\text{x}$ [$^\circ$] & R$_\text{y}$ [$^\circ$] & R$_\text{z}$ [$^\circ$] \\\hline
          Calibrated & $0.014$ & $0.115$ & $0.024$ \\
          0.01$^\circ$ Drift & $0.021$ & $0.126$ & $0.032$ \\
          & T$_\text{x}$ [mm] & T$_\text{y}$ [mm] & T$_\text{z}$ [mm] \\\hline
          Calibrated & $0.98$ & $2.53$ & $7.06$ \\
          0.01$^\circ$ Drift & $1.33$ & $4.35$ & $9.42$ \\
    \end{tabular}
\subcaption{Geometric Precision (lower is better)}
\label{tab:geometric_precision_man}
\end{subtable}\vspace{1em}
    \begin{subtable}[h!]{0.5\textwidth}
    \centering
    \begin{tabular}{c|ccc}
         Data & KO-I [px] & VOF-I [px] & DC-I [m] \\\hline
         Calibrated & $-$0.002 & $-$0.096 & 0.166\\
         0.01$^\circ$ Drift & 0.029 & $-$0.076 & 0.170 \\
    \end{tabular}
\subcaption{Stereo Metrics (lower is better, negative means improvement over the reference parameters)}
\label{tab:stereo_metrics_man}
\end{subtable}
\caption{TESO performance on the MAN TruckScenes dataset, visualised as geometric precision \ref{tab:geometric_precision_man} and stereo metrics \ref{tab:stereo_metrics_man}. }
\label{tab:man_results}
\hfill
\end{table}

\subsection{Global Optimization on CARLA--FlowGuided}
 This experiment shows that the proposed kernelized error is as robust and effective as published learning-based methods. To facilitate the comparison, we utilize the CARLA-FlowGuided dataset~\cite{kumar2024}, on which numerous results have been published. This dataset was specifically developed with end-to-end learning methods in mind. Rather than providing the full sequences from each ``drive'', the dataset contains only a~discrete sample of recorded stereo image pairs. Approximately 90\,\% of the data are in the training part and the rest is in the validation part. We evaluated the TESO loss function solely on the validation part of the dataset. That is, no stage of our method has seen any frame from the training part of the dataset.

Since the available part of the dataset does not contain any sequence, we cannot employ the online stochastic optimization of TESO. Therefore, we evaluate the performance of the kernelized error (see \cref{eq:kernel_based_error}) alone. To this end, we use a~differential evolution (DE) optimization~\cite{Storn1997} on the essential matrix manifold, \cref{eq:emm_map}. We do so for seven iterations with the annealing of the $\sigma$ parameter from $0.02$ (halved with each iteration). The idea behind this approach is to transition from a~coarse estimate of the parameters to more precise ones as $\sigma$ decreases.

The results of our extrinsic calibration using DE are compared in \cref{tab:carlafg_results} with those of the best three published SOTA methods~\cite{kumar2024, rockwell2024, gong2025}. Our approach achieves competitive results. This suggests that the proposed kernel correlation error \cref{eq:kernel_based_error} is fully adequate for the task. Interestingly, unlike the results on the MAN TruckScenes dataset (see \cref{tab:man_results}), the geometric precision on the CARLA--FlowGuided dataset is also significantly better. This appears to be caused by the properties of CARLA--FlowGuided cameras, specifically a~higher number of pixels (2560 vs. 1928),  despite having a~much narrower field of view (45$^\circ$ vs. 120$^\circ$), no vergence and shorter baseline.

\begin{table}[]
    \begin{subtable}[h!]{0.5\textwidth}
    \centering
    \begin{tabular}{c|cccc}
          Method & R$_\text{x}$ [$^\circ$] & R$_\text{y}$ [$^\circ$] & R$_\text{z}$ [$^\circ$]  & T [$^\circ$] \\\hline
          \citet{kumar2024} & 0.03 & 0.23 & 0.11 & --- \\
          \citet{rockwell2024} & 0.007 &  0.153 & 0.017 & 2.73 \\
          \citet{gong2025} & $\textbf{0.003}$ & $0.077$ & $\textbf{0.006}$ & $\textbf{0.86}$\\
          Ours (DE) & $0.007$ & $\textbf{0.027}$ & $0.012$ & $1.34$\\
    \end{tabular}
\subcaption{Geometric Precision (lower is better)}
\label{tab:geometric_precision_carlafg}
\end{subtable}\vspace{1em}
    \begin{subtable}[h!]{0.5\textwidth}
    \centering
    \begin{tabular}{cccc}
         Method & KO [px] & VOF [px] & DC [m] \\\hline
           GT & 0.47 & 0.51 & --- \\
           \citet{kumar2024} & 0.64 & 1.00 & 5.6 \\
          \citet{rockwell2024} & 0.56 & 0.57 & 4.9 \\
           \citet{gong2025} & \textbf{0.51} & \textbf{0.52} & \textbf{3.2} \\
           Ours (DE) & \textbf{0.51} & 0.55 & 4.6 \\
    \end{tabular}
\subcaption{Stereo Metrics (lower is better)}
\label{tab:stereo_metrics_carlafg}
\end{subtable}
\caption{Results of our kernelized error \cref{eq:kernel_based_error} optimization using differential evolution (DE) on CARLA--FlowGuided dataset and current SotA methods on this dataset (taken from~\cite{kumar2024, gong2025}). Our recalibration shows competitive results with the best end-to-end learning method~\cite{gong2025}. This suggests very good robustness of our kernel-based error to outlier matches, without any need for training on new datasets. }
\label{tab:carlafg_results}
\hfill
\end{table} \section{Conclusion}
\label{sec:conclusion}
In this work, we propose TESO, an online stochastic optimization method for temporal tracking of the essential matrix. In contrast to other recent works, it does not require any data-specific training and has a~small computational overhead. As it uses an inherently robust kernelized epipolar error, it works with potentially low-quality, easy-to-extract tentative matches without requiring explicit outlier rejection. We have conducted a~thorough analysis of the approach using four datasets, demonstrating its performance under various conditions. In the KITTI dataset, we have identified an inconsistency in the offline calibration parameters across several stereo pairs, which corroborates the findings of other authors~\cite{Cvisic21}. On the MAN TruckScenes dataset, we showed good calibration and drift tracking results of TESO under highly realistic conditions (variable weather, repetitive highway scenes, wide baseline, and low pixel angular resolution). Using the CARLA--FlowGuided dataset, we demonstrated that the proposed kernelized loss function achieved precision comparable to that of end-to-end learning methods across several metrics, without any use of the training part of the dataset. We have also introduced a~CARLA--Drift dataset that contains a~known, fast drift in the camera's orientation, allowing us to evaluate TESO on sequences with ground truth. It will be made publicly available along with the TESO inference code. It is suitable for implementation on low-cost hardware or, potentially, for ASICs in integrated sensor systems with large baselines. 
\section*{Acknowledgments}
This work was supported by the Czech Technical
University in Prague [Grant~SGS24/096/OHK3/2T/13], by the NII International Internship Program, and in part by the
Technology Agency of the Czech Republic under the National Competence
Centres II Programme [Project \#~TN02000054 `Bo\v zek Vehicle
Engineering NCC-II' (BOVENAC)]. An earlier phase of this work was supported by the OP VVV MEYS project ‘Research Center for Informatics’ [Grant CZ.02.1.01/0.0/0.0/16019/0000765].

{
    \small
    \bibliographystyle{ieeenat_fullname}
    \bibliography{main}
}

\onecolumn

\maketitlesupplementary

\setcounter{section}{0}
\def\thesection{A.\arabic{section}}

\section{An ablation study: SIFT vs. SuperGlue and Kernel correlation vs. Non-robust loss}\label{seq:ablat}\label{seq_a:ablation}
This experiment illuminates the robustness introduced by the kernel correlation (KC) principle, demonstrating that the selection of the keypoint detector, feature extractor, and matcher becomes non-critical once the kernelized loss function is used. 

We have evaluated TESO performance on three different combinations of keypoints and loss functions:
\begin{enumerate}
    \item SIFT, w/ KC, 5-NN (standard TESO, see Methods section of the paper) with
\begin{equation}\label{eq:sift_robust}
    \mathcal{L}(\mathbf{\theta}\,|\,\mathbf{X}, \mathbf{Y}) = -\sum_{\mathbf{x}\in \mathbf{X}}\,\sum_{\mathbf{y}\in \text{NN}^1(\mathbf{x})}\text{exp}\left[-\frac{\left(\mathbf{y}^\top \mathbf{E}(\mathbf{\theta})\mathbf{x}\right)^2}{2\sigma^2}\right] - 
     \sum_{\mathbf{y}\in \mathbf{Y}}\sum_{\mathbf{x}\in \text{NN}^0(\mathbf{y})}\text{exp}\left[-\frac{\left(\mathbf{y}^\top \mathbf{E}(\mathbf{\theta})\mathbf{x}\right)^2}{2\sigma^2}\right],
\end{equation}
\item SuperGlue~\cite{sarlin2020} matches $(\mathbf{x, y})\in \text{SG}$, w/o KC, i.\,e.:
\begin{equation}\label{eq:sg_nonrobust}
    \mathcal{L}(\mathbf{\theta}\,|\,\text{SG}) = -\sum_{(\mathbf{x, y})\in \text{SG}}\left(\mathbf{y}^\top \mathbf{E}(\mathbf{\theta})\mathbf{x}\right)^2,
\end{equation}
\item SuperGlue matches $(\mathbf{x, y})\in \text{SG}$, w/ KC, i.\,e.:
\begin{equation}\label{eq:sg_robust}
    \mathcal{L}(\mathbf{\theta}\,|\,\text{SG}) = -\sum_{(\mathbf{x, y})\in \text{SG}}\text{exp}\left[-\frac{\left(\mathbf{y}^\top \mathbf{E}(\mathbf{\theta})\mathbf{x}\right)^2}{2\sigma^2}\right].
\end{equation}
\end{enumerate}

SuperGlue provides one-to-one matches with a~specific confidence level. If the matcher works correctly, matches with higher confidence should have a~higher probability of being inliers of the epipolar geometry. TESO tracking without a~robust loss function (\cref{eq:sg_nonrobust}) should therefore show better performance on keypoints with higher confidence levels. The kernel correlation (\cref{eq:sg_robust}) should robustify the results, ensuring that the tracking performance is less dependent on the quality of the matches.

\begin{figure}[t]
    \centering
    \includegraphics[width=0.495\linewidth]{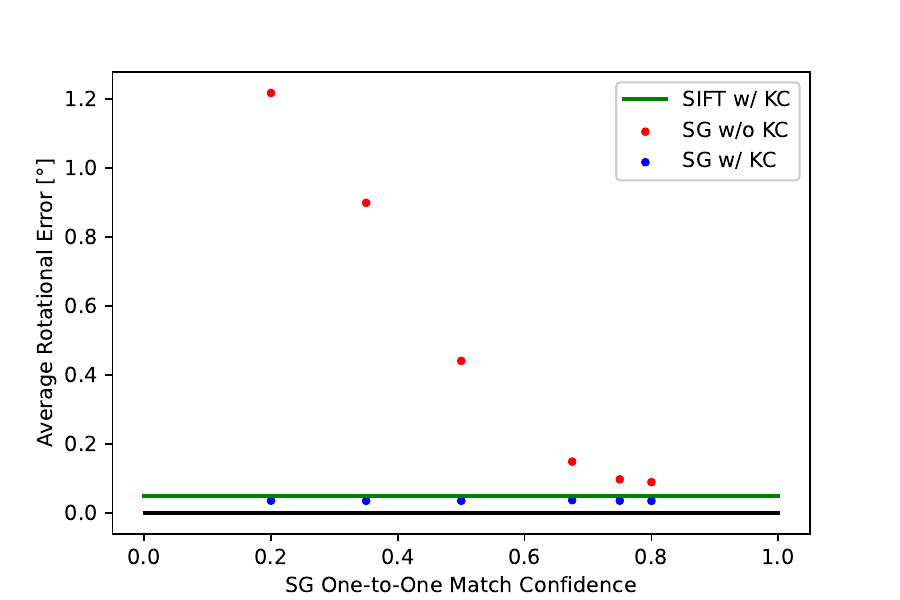}
    \includegraphics[width=0.495\linewidth]{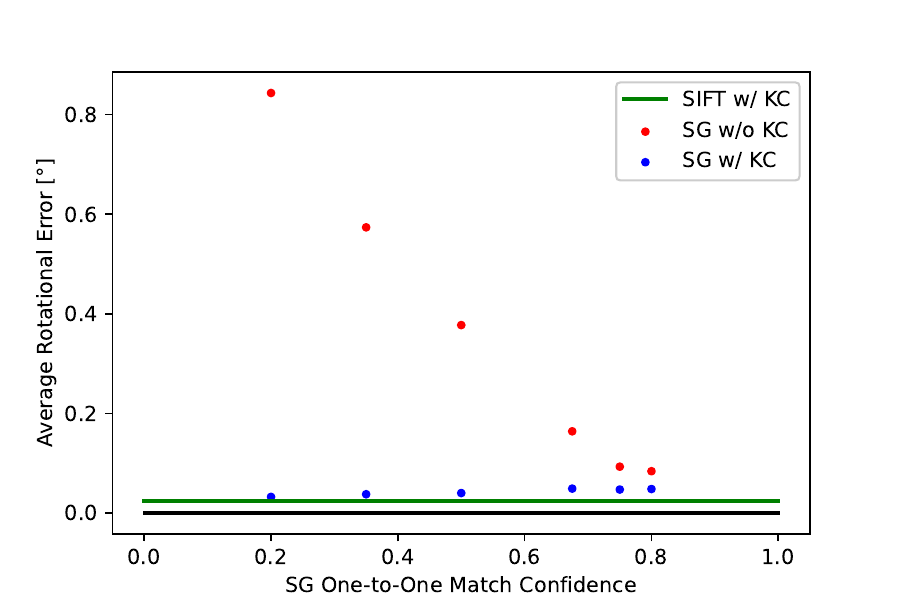}
    \caption{TESO performance visualized as an average rotational error on two sequences from the MAN TruckScenes dataset using different matching algorithms and loss functions. It is evident that with a~higher confidence level in SuperGlue matches and a~non-robust loss function (represented by the red points), the precision increases. However, using our proposed kernelized loss function with SuperGlue matches (blue points) renders the confidence hyper-parameter unimportant. The difference between SIFT (with 5-NN matching) and SuperGlue, using a~kernelized loss function, is also very small, suggesting that the robust loss function renders the selection of keypoint detector and feature extractor uncritical. In all cases, the proposed robust loss function outperforms the non-robust variant.}
    \label{fig:are_kernel_or_not}
\end{figure}

In \cref{fig:are_kernel_or_not}, we present the average rotational error (ARE) of TESO tracking on two sequences from the MAN TruckScenes dataset (around 160 frames each) under various scenarios. In ARE, lower values indicate better performance, i.e., TESO tracking is closer to the reference parameters (black line). The x-axis represents the minimum confidence level of the SuperGlue matches. SIFT has a~constant value (green line) as it does not depend on the SG confidence. For the non-robust version of the loss (\cref{eq:sg_nonrobust}, red points), as the confidence of SuperGlue matches increases, the average rotational error decreases. The robust kernel correlation (\cref{eq:sg_robust}) on SuperGlue features (blue points) renders the confidence parameter unimportant and is always more precise than the non-robust variant (\cref{eq:sg_nonrobust}). When TESO uses SIFT (with 5-NN matches, as proposed in the paper), it achieves slightly worse precision on one sequence (left plot) and slightly better precision on the other (right plot) than SuperGlue matches with a~kernelized loss function (blue points).

These results suggest that selecting the keypoint detector, feature extractor, and matcher is not crucial for the kernelized loss function.

\section{Bias and latency evaluation}
In this experiment, we demonstrate that the TESO tracker is unbiased and exhibits low latency in tracking. 

Instead of examining the mean absolute error of the tracker in rotation (as in Section 5.3 of the paper), we evaluate the actual average tracked parameters on 149 calibrated sequences from the MAN TruckScenes dataset (with no simulated calibration drift). Here, we artificially extend the sequences by repeating them periodically ten times (this yields approximately 1600 frames per sequence). TESO achieved the following mean parameters over all sequences (with standard deviation):
\begin{center}
\begin{tabular}{|c|c|c|}\hline
    R$_\text{x}$ &R$_\text{y}$ & R$_\text{z}$\\\hline
     $-0.0009$ ($\pm0.017$) & $-0.0682$ ($\pm0.110$) & $0.0143$ ($\pm0.023$) \\\hline
\end{tabular}
\end{center}
As the parameter deviation from zero is not statistically significant, it suggests that the tracker is unbiased.

\begin{figure}[t]
    \centering
    \includegraphics[width=0.495\linewidth]{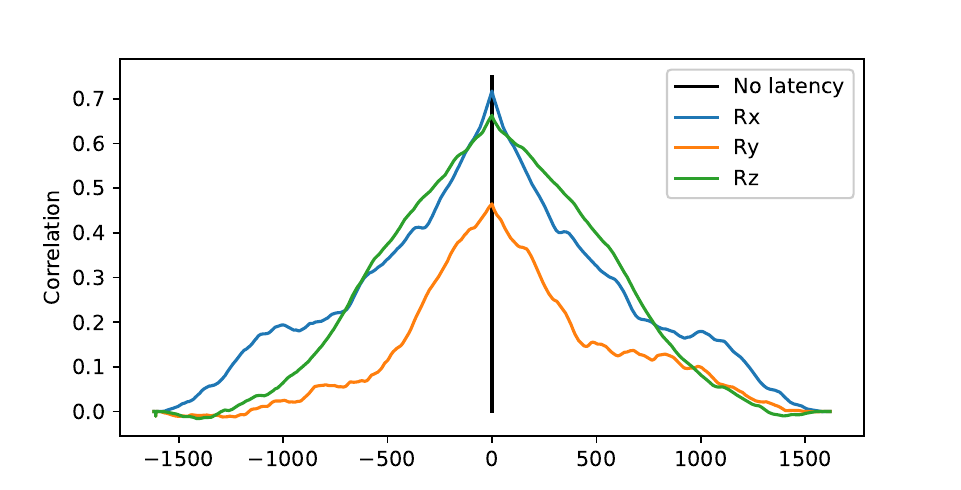}
    \caption{Latency evaluation on MAN TruckScenes dataset. It is estimated as a~discrete cross-correlation between the TESO tracking sequence and the cumulative drift sequence for shifts in the range $(-1600, 1600)$. The maxima in all three  degrees of freedom are at zero, which suggests the latency is less than one frame.}
    \label{fig:latency}
\end{figure}

On the drifted sequences (cumulated drift of $\pm0.01^\circ$/DoF/frame), we evaluated the cross-correlation profile between the cumulated drift sequence and the TESO tracking sequence. \cref{fig:latency} shows that the maximum correlation in all three  degrees of freedom is zero (the sequences are discrete, with a~unit of one frame). Although we have anticipated a~more substantial latency (at least a~frame), this result suggests that the mean latency is less than a~frame in the TESO reaction to the cumulative drift. It is not possible to find a~sub-frame latency estimate with this correlation method.
\section{Memory and time efficiency of TESO}
\subsection*{Memory}
Throughout the stochastic optimization, TESO needs to temporally store the following variables:
\begin{itemize}
    \item a~vector of filtered gradients $\mathbf{g}\in \mathbb{R}^5$,
    \item a~vector of filtered variance of gradients $\mathbf{v}\in \mathbb{R}^5$
    \item a~vector of filtered diagonal of the Hessian matrix $\mathbf{h}\in \mathbb{R}^5$,
    \item memory size of the filter $\mathbf{m}\in \mathbb{R}^5$ and
    \item orthogonal matrices that make up the essential manifold $\mathbf{U}, \mathbf{V}\in \mathbb{R}^{3\times3}$.
\end{itemize}
That is, a~total of 38 parameters.
\subsection*{Time}
We run a~Python implementation of TESO on a~desktop CPU AMD Ryzen 5 9600X. On one sequence from the CARLA--Drift dataset (with 1024$\times$512\,px resolution and extracting at most 1000 keypoints per frame), we achieved the following per-frame time efficiency:
\begin{center}
\begin{tabular}{|l|c|}\hline
    & Time [ms] \\ \hline
   Keypoint detection and feature extraction~\cite{lowe1999,opencv_library} & 70.16 \\ \hline
   Nearest neighbors search~\cite{douze2024} & 8.50 \\ \hline
   Gradients and the diagonal of the Hessian matrix estimation  & 7.24 \\ \hline
   Filter and manifold update  & 0.11 \\ \hline
   \textbf{Total}  & \textbf{86.01} \\ \hline
\end{tabular}
\end{center}
The most time-consuming part, as expected, is keypoint detection and feature extraction. As shown in \cref{seq:ablat}, the use of a~robust kernelized loss function renders the choice of a~keypoint detector uncritical, allowing for the use of a~more time-efficient variant. Our CPU implementation of TESO achieves a~runtime similar to that of the fastest SotA method~\cite{gong2025} (79\,ms vs.~ours 86\,ms), whose matcher (41\,ms vs.~ours 79\,ms) runs on a~GPU and utilizes data-driven training.

\section{Keypoint detector selection}\label{seq_a:keypoint_detector}
The keypoint detector is a~building block of the TESO pipeline that may influence its precision and robustness. In this experiment, we have tested several kinds of keypoint detectors and feature extractors. You can see the TESO tracking average rotational error (ARE; lower is better) across three tested datasets in the table below. We did not find statistically significant differences between SIFT and BRISK across the datasets and all sequences; both performed best. This is consistent with the related work~\cite{Tareen2018}. If a~faster execution is needed, ORB can be used, but a~drop in precision should be expected.

\begin{center}
\begin{tabular}{|c|c|c|c|c|}\hline
    &\multicolumn{4}{|c|}{Average Rotational Error $[^\circ]$}\\
    &SIFT & BRISK & ORB & STAR + BRIEF64\\\hline
    CARLA--Drift & $\textbf{0.022}$ & $\textbf{0.022}$ & $0.037$ & $0.027$ \\\hline
    KITTI, 00-01, \cite{Cvisic21} & $\textbf{0.014}$ & $\textbf{0.014}$ & $0.025$ & $0.020$ \\\hline
    MAN, 0.01$^\circ$ Drift & $\textbf{0.049}$ & $0.052$ & $0.062$ & $0.057$ \\\hline
\end{tabular}
\end{center}
\section{Hyper-parameters selection}
Our method has two hyper-parameters -- $\sigma$ and $k$. The parameter $\sigma$ represents a~trade-off between the basin of attraction width and the variance of the tracked parameters; see the table below. In calibrated data (\textit{CARLA–Drift, Calibrated} row), $\sigma$ increases the TESO variance, but it also helps to track the drift (\textit{CARLA--Drift, 0.02$^\circ$ Drift} row). This depends on the expected decalibration speed for the setup and the sensor resolution. We have selected $\sigma=0.001$ for CARLA--Drift and MAN TruckScenes dataset, which seems to be a~good trade-off between the basin of attraction and process variance. For KITTI, we selected a~slightly lower value of $\sigma=0.00075$ due to a~lower angular resolution of the pixels ($0.05^\circ/\text{px}$) compared to CARLA ($0.068^\circ/\text{px}$).

\begin{center}
\begin{tabular}{|c|c|c|c|c|c|}\hline
    &\multicolumn{5}{|c|}{Average Rotational Error $[^\circ]$}\\
    &$\sigma=0.00025$ & $\sigma=0.0005$ & $\sigma=0.001$ & $\sigma=0.002$ & $\sigma=0.004$\\\hline
    CARLA--Drift, Calibrated & $0.004$ &$0.005$ & $0.007$ & $0.009$ & $0.015$ \\\hline
    CARLA--Drift, 0.02$^\circ$ Drift & $0.051$ &$0.033$ & $0.029$ & $0.029$ & $0.032$\\\hline
\end{tabular}
\end{center}

The parameter $k$ in kNN further robustifies the method for challenging (repetitive) scenes. In high-quality data, $k=1$ will be the most precise; we have chosen $k=5$, which is more conservative (a closer bound on the theoretical kernel correlation~\cite{tsin04}), but may reduce geometric precision.

\section{Analysis of TESO results on the MAN TruckScenes dataset}\label{seq_a:man}
To further analyze the quality of TESO tracking in challenging scenarios, we present results from the MAN dataset, broken down by daytime, location, and weather for each sequence, in \cref{tab:man_full}. In both calibrated and drifted sequences, tracking is statistically significantly worse at night, in fog, or in tunnels (i.e., weather is `Other'). This is expected behavior, as universal keypoint detectors (such as SIFT~\cite{lowe1999}) struggle under low-light conditions. Such scenarios likely require specialized, pre-trained keypoint detectors. A~universal SIFT keypoint detector appears sufficient for any driving location, including rainy weather.
\begin{table}[]
     \begin{subtable}{\textwidth}
    \centering
     \scriptsize{
    \begin{tabular}{ccc|cccccc}
         &Data & Count &R$_\text{x}$ [$^\circ$] & R$_\text{y}$ [$^\circ$] & R$_\text{z}$ [$^\circ$] &T$_\text{x}$ [mm] & T$_\text{y}$ [mm] & T$_\text{z}$ [mm] \\\hline
         &All & 149 & $0.014\,(\pm 0.01)$ & $0.115\,(\pm 0.07)$ & $0.024\,(\pm0.02)$ & $1.0\,(\pm 0.7)$ & $2.5\,(\pm 2.7)$ & $7.1\,(\pm5.3)$ \\\hline
         \multirow{3}{*}{\rotatebox[origin=c]{90}{Daytime}} & Morning & 45 & $0.013$ & $0.119$ & $0.020 $ & $1.1$ & $2.7$ & $8.1$\\
          & Noon & 97 & $0.016$ & $0.106$ & $0.023$ & $0.9$ & $2.4$ & $6.2$ \\
          & \textcolor{red}{Evening} & 7 & $0.015$ & \textcolor{red}{$0.208$} & \textcolor{red}{$0.054$} & $1.7$ & $3.8$ & $11.7$ \\\hline
         \multirow{6}{*}{\rotatebox[origin=c]{90}{Location}} & Terminal & 12 & $0.024$ & $0.057$ & $0.034 $ & $1.5$ & $5.1$ & $10.6$\\
          & Highway & 103 & $0.014$ & $0.131$ & $0.024$ & $1.0$ & $2.4$ & $7.2$ \\
          & Residential & 5 & $0.016$ & $0.084$ & $0.018$ & $0.8$ & $3.3$ & $5.7$ \\
          & City & 7 & $0.015$ & $0.054$ & $0.027$ & $0.4$ & $1.6$ & $3.1$ \\
          & Parking & 2 & $0.006$ & $0.097$ & $0.015$ & $0.8$ & $0.6$ & $5.8$ \\
          & Rural & 20 & $0.011$ & $0.097$ & $0.017$ & $0.9$ & $1.9$ & $6.2$ \\\hline
         \multirow{6}{*}{\rotatebox[origin=c]{90}{Weather}} & Clear & 64 & $0.015$ & $0.120$ & $0.026 $ & $1.0$ & $2.6$ & $6.9$\\
          & Overcast & 52 & $0.010$ & $0.100$ & $0.014$ & $0.9$ & $1.6$ & $6.6$ \\
          & Rain & 30 & $0.018$ & $0.129$ & $0.032$ & $1.1$ & $3.7$ & $7.8$ \\
          & \textcolor{red}{Fog} & 1 & \textcolor{red}{$0.027$} & $0.078$ & $0.037$ & $0.6$ & \textcolor{red}{$9.4$} & $5.3$ \\
          & \textcolor{red}{Other} & 2 & $0.013$ & $0.129$ & \textcolor{red}{$0.051$} & \textcolor{red}{$1.9$} & $4.9$ & \textcolor{red}{$13.0$} \\\hline
    \end{tabular}}
    \caption{Calibrated}
    \end{subtable}
    \par\vspace{2ex}
    \begin{subtable}{\textwidth}
    \centering
    \scriptsize{
    \begin{tabular}{ccc|cccccc}
         &Data & Count &R$_\text{x}$ [$^\circ$] & R$_\text{y}$ [$^\circ$] & R$_\text{z}$ [$^\circ$] &T$_\text{x}$ [mm] & T$_\text{y}$ [mm] & T$_\text{z}$ [mm] \\\hline
         &All & 149 & $0.021\,(\pm 0.01)$ & $0.126\,(\pm 0.07)$ & $0.032\,(\pm0.02)$ & $1.3\,(\pm 0.8)$ & $4.4\,(\pm 2.9)$ & $9.4\,(\pm5.6)$ \\\hline
         \multirow{3}{*}{\rotatebox[origin=c]{90}{Daytime}} & Morning & 45 & $0.019$ & $0.132$ & $0.027 $ & $1.4$ & $4.1$ & $10.1$\\
          & Noon & 97 & $0.021$ & $0.119$ & $0.031$ & $1.3$ & $4.4$ & $9.1$ \\
          & \textcolor{red}{Evening} & 7 & $0.024$ & $0.191$ & \textcolor{red}{$0.060$} & $1.4$ & $5.1$ & $9.8$ \\\hline
         \multirow{6}{*}{\rotatebox[origin=c]{90}{Location}} & Terminal & 12 & $0.027$ & $0.068$ & $0.039 $ & $1.5$ & $5.2$ & $10.4$\\
          & Highway & 103 & $0.021$ & $0.142$ & $0.032$ & $1.4$ & $4.4$ & $9.7$ \\
          & Residential & 5 & $0.021$ & $0.100$ & $0.026$ & $1.0$ & $5.0$ & $6.8$ \\
          & City & 7 & $0.019$ & $0.079$ & $0.034$ & $0.8$ & $2.9$ & $5.7$ \\
          & Parking & 2 & $0.012$ & $0.117$ & $0.018$ & $0.8$ & $4.8$ & $5.7$ \\
          & Rural & 20 & $0.017$ & $0.106$ & $0.025$ & $1.4$ & $3.9$ & $9.8$ \\\hline
         \multirow{6}{*}{\rotatebox[origin=c]{90}{Weather}} & Clear & 64 & $0.021$ & $0.126$ & $0.033 $ & $1.1$ & $3.9$ & $8.0$\\
          & Overcast & 52 & $0.016$ & $0.109$ & $0.021$ & $1.5$ & $3.9$ & $10.6$ \\
          & Rain & 30 & $0.027$ & $0.145$ & $0.043$ & $1.4$ & $5.9$ & $9.9$ \\
          & \textcolor{red}{Fog} & 1 & \textcolor{red}{$0.038$} & \textcolor{red}{$0.441$} & \textcolor{red}{$0.058$} & $1.6$ & \textcolor{red}{$8.0$} & $12.1$ \\
          & \textcolor{red}{Other} & 2 & $0.020$ & $0.135$ & \textcolor{red}{$0.058$} & \textcolor{red}{$2.7$} & $5.1$ & \textcolor{red}{$17.8$} \\\hline
    \end{tabular}}
    \caption{0.01$^\circ$ calibration drift}
    \end{subtable}
    \caption{Full results of TESO on the MAN Dataset based on the time of the day, location, and weather of each sequence. It shows results for sequences with correct calibration (a) and for sequences with synthetic calibration drift in all rotational DoFs (b). Red shows a~statistically significantly worse precision than the tracking results over all sequences.}
    \label{tab:man_full}
\end{table}
\section{Translation tracking precision}
\begin{figure}[t]
    \centering
    \begin{subfigure}[t]{0.35\linewidth}
        \centering
        \includegraphics[width=\linewidth]{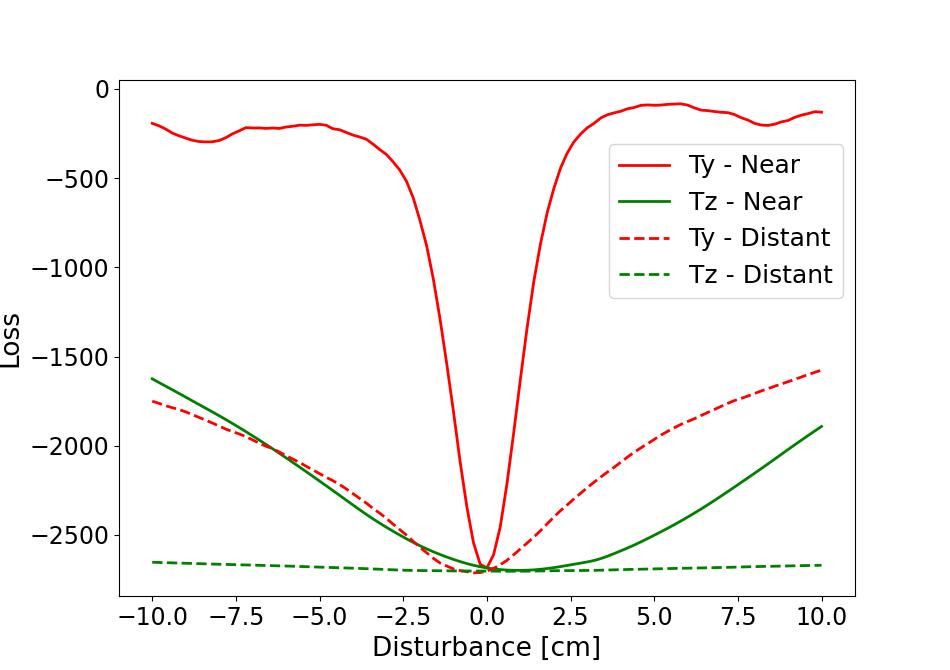}
        \caption{Near vs. Distant Scene}\label{fig:near_vs_distant}
    \end{subfigure}\begin{subfigure}[t]{0.35\linewidth}
        \centering
        \includegraphics[width=\linewidth]{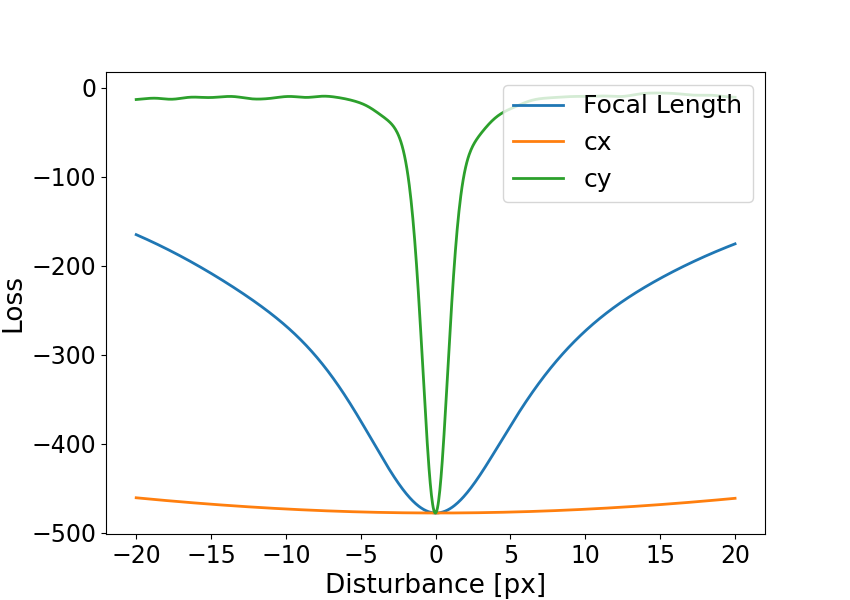}
        \caption{Intrinsics (non-parallel stereo)}\label{fig:intri}
    \end{subfigure}
    \caption{Kernel corelation loss evaluations. Narrower is better.\vspace{-1.5em}}
\end{figure}

Throughout this work, we have focused on rotation precision only. This is because the translation is not as observable as the rotations in automotive scenarios, given the very distant scene. Shown in \cref{fig:near_vs_distant} is the kernel correlation loss evaluation: Narrow curves indicate a~higher sensitivity of the loss function (y-axis) to decalibration (x-axis). Except in the Y translation in near scenes, this shows that achieving centimeter-level precision with feature-based optimization methods in typical driving scenes is difficult (as opposed to tight calibration rooms). Position calibrations obtained from vehicle drawings are inherently more accurate than those from online calibration methods.

\section{Intrinsics calibration}
Our experiments show that the focal length and the vertical position of the principal point are observable; see \cref{fig:intri}. Note that the sensitivity of the focal length is quite high, five pixels are less than 1\,\% of the image size.

However, adding it to TESO is not straightforward, as these are not manifold parameters.  It would require changing the optimization scheme used in TESO. See also the last paragraph of Section 5.3 of the paper.

\end{document}